\documentclass[preprint,5p,times,twocolumn]{elsarticle}
\usepackage{amssymb}
\usepackage{amsmath}
\usepackage{hyperref}
\usepackage{url}
\usepackage{booktabs} 
\usepackage{multirow}
\journal{Nuclear Physics B}

\begin{document}

\begin{frontmatter}



\title{TransMA: an explainable multi-modal deep learning model for predicting properties of ionizable lipid nanoparticles in mRNA delivery} %% Article title


\author[label1,label4]{Kun Wu\fnref{fn1}} 
\ead{wuk@sari.ac.cn}
\author[label2]{Zixu Wang\fnref{fn1}} 
\ead{s2230167@u.tsukuba.ac.jp}
\author[label3,label5]{Xiulong Yang} 
\ead{yangxiulong@ccnu.edu.cn}
\author[label2]{Yangyang Chen} 
\ead{chen.yangyang.xp@alumni.tsukuba.ac.jp}
\author[label1,label4]{Zhenqi Han} 
\ead{hanzq@sari.ac.cn}
\author[label1,label4]{Jialu Zhang} 
\ead{zhangjl@sari.ac.cn}
\author[label1,label4]{Lizhuang Liu\corref{cor1}} 
\ead{liulz@sari.ac.cn}
\affiliation[label1]{organization={Shanghai Advanced Research Institute, Chinese Academy of Sciences},
            city={Shanghai},
            postcode={201210}, 
            country={China}}
\affiliation[label2]{organization={Department of Computer Science, University of Tsukuba},
            city={Tsukuba},
            postcode={3058577}, 
            country={Japan}}
\affiliation[label3]{organization={Hubei Provincial Key Laboratory of Artificial Intelligence and Smart Learning, Central China Normal University},
            city={Wuhan},
            postcode={430079}, 
            country={China}}
\affiliation[label4]{organization={University of Chinese Academy of Sciences},
            city={Beijing},
            postcode={100049}, 
            country={China}}
\affiliation[label5]{organization={School of Computer Science, Central China Normal University},
            city={Wuhan},
            postcode={430079}, 
            country={China}}        
\cortext[cor1]{Corresponding author.}
\fntext[co-first]{These authors contributed equally to this work.}
%% Abstract
\begin{abstract}
As the primary mRNA delivery vehicles, ionizable lipid nanoparticles (LNPs) exhibit excellent safety, high transfection efficiency, and strong immune response induction. However, the screening process for LNPs is time-consuming and costly. To expedite the identification of high-transfection-efficiency mRNA drug delivery systems, we propose an explainable LNPs transfection efficiency prediction model, called TransMA. TransMA employs a multi-modal molecular structure fusion architecture, wherein the fine-grained atomic spatial relationship extractor named molecule 3D Transformer captures three-dimensional spatial features of the molecule, and the coarse-grained atomic sequence extractor named molecule Mamba captures one-dimensional molecular features. We design the mol-attention mechanism block, enabling it to align coarse and fine-grained atomic features and captures relationships between atomic spatial and sequential structures. TransMA achieves state-of-the-art performance in predicting transfection efficiency using the scaffold and cliff data splitting methods on the current largest LNPs dataset, including Hela and RAW cell lines. Moreover, we find that TransMA captures the relationship between subtle structural changes and significant transfection efficiency variations, providing valuable insights for LNPs design. Additionally, TransMA's predictions on external transfection efficiency data maintain a consistent order with actual transfection efficiencies, demonstrating its robust generalization capability. The code, model and data are made publicly available at \url{https://github.com/wklix/TransMA/tree/master}. We hope that high-accuracy transfection prediction models in the future can aid in LNPs design and initial screening, thereby assisting in accelerating the mRNA design process.
\end{abstract}


\begin{highlights}
\item An explainable LNPs transfection efficiency prediction model called TransMA. 
\item TransMA employs a multi-modal molecular structure fusion architecture.
\item TransMA reveals the atomic-level structure-transfection relationships.
\item TransMA achieves state-of-the-art performance on the largest current LNPs dataset. 
\end{highlights}

%% Keywords

\end{frontmatter}
\section{Introduction}
mRNA-based technologies hold promise for therapeutic applications in fields such as viral vaccines, protein replacement therapies, cancer immunotherapies, genome editing\cite{1,2,3,4,5,6}, and have the potential to reshape the landscape of life science research and medicine\cite{7,8,9}. However, achieving targeted delivery and intracellular release from endosomes remains challenging for mRNA delivery systems, emphasizing the critical demand for safe and effective mRNA delivery materials. Specifically, lipid nanoparticles (LNPs) have undergone extensive investigation and have successfully transitioned into clinical use for the delivery of mRNA\cite{10,11,12}. For example, mRNA1273 and BNT162b21 utilize lipid nanoparticles for the delivery of antigen mRNA\cite{13,14}. LNPs consist of four components: ionizable lipid, phospholipid, cholesterol, and PEGylated lipid. Among these, the ionizable lipid component holds the highest molar ratio, which determines the formulation's delivery efficiency and stability, serving as the core structure of LNPs\cite{15}. Therefore, the key to selecting for efficient LNPs lies in the selection of the ionizable lipid\cite{16,17}.

Although component chemistry methods based on three-component reactions (3-CR)\cite{20,21} and four-component reactions (4-CR)\cite{47} can synthesize a number of ionizable lipids, manually testing the transfection efficiency of each synthesized lipid is time-consuming and costly\cite{18,19}. Several works have shown that artificial intelligence represented by machine learning and deep learning can achieve automatic prediction of LNPs transfection efficiency. For instance, Ding et al.\cite{9} employed four machine learning methods—support vector machine, random forest, eXtreme gradient boosting, and multilayer perceptron—to classify the transfection efficiency of 572 lipid nanoparticles (LNPs) into two categories, achieving a classification accuracy of 98$\%$. AGILE \cite{22} is a prediction platform for LNPs transfection efficiency based on graph convolutional neural networks and pre-training, achieving a mean squared error (MSE) of approximately 6 in predicting LNPs transfection efficiency. TransLNP\cite{31} which is a model based on Transformer architecture and data balancing achieved a MSE of approximately 5 on the AGILE dataset.

Despite significant progress in previous works, it must be acknowledged that there are notable limitations in achieving automatic prediction of LNPs transfection efficiency based on machine learning and deep learning. These limitations hinder the accuracy and generalizability of the models. These limitations include:(1)\textbf{The lack of multi-modal information interaction.} Previous works only extract single-modal information from molecules, which limits their ability to capture Quantitative Structure-Activity Relationships (QSAR) between ionizable lipids and transfection efficiency. For instance, chemberta \cite{36} relies on one-dimensional representations to predict molecular properties. Integrating multi-modal information from molecules is crucial for further improving prediction accuracy. (2)\textbf{Previous works lacks interpretability.} Although previous works can predict transfection efficiency accurately, it is unclear which atoms play a key role in this process. This lack of interpretability restricts deep understanding of the prediction process and hampers efforts to further optimize the model for improved accuracy. Therefore, enhancing model interpretability while maintaining prediction accuracy poses a significant challenge in current research.  (3)\textbf{Limited attention to the impact of transfection cliffs.} Pairs of molecules that are structurally very similar but have significantly different transfection efficiencies—known as transfection cliffs—capture the knowledge hidden in the elusive structure-property relationships but have received limited attention.

To address the challenges, we introduce an explainable LNPs transfection efficiency prediction model called TransMA, which adopts a multi-modal molecular structure fusion architecture. TransMA integrates 3D geometric information and 1D atomic sequence information of molecules to predict LNPs transfection efficiency. A self-attention mechanism SE(3) Transformer architecture named molecule 3D Transformer is employed to extract 3D geometric information. Molecule 3D Transformer is pre-trained by reconstructing atomic 3D coordinates and masked atom prediction. To extract atomic sequence information, the state space model designated as molecule Mamba is presented to obtain coarse-grained atomic sequence information. By constructing multi-level molecular structure data pairs, the model jointly learns the multi-dimensional structural features of molecules. Specifically, we design the mol-attention mechanism block to align and concatenate the fusion features. The fusion features include atomic sequence distribution information, atomic coordinates information, relative atomic positions information, and types of bonds between atoms information. Compared to state-of-the-art molecular graph convolutional networks and Transformer models, TransMA demonstrates the best performance, reducing MSE by 43$\%$ compared to the baseline model AGILE and by 35$\%$ compared to TransLNP.  Additionally, TransMA possesses interpretability to better understand the mapping relationship between molecular structure and biochemical properties\cite{23,24}. The mol-attention mechanism block not only integrates multi-modal molecular features but also reveal critical sites that influence transfection efficiency. Transfection cliffs represent pairs of molecules that are structurally similar but exhibit substantial differences in transfection efficiency. Extracting these transfection cliff data can help us understand the elusive transfection efficiency. A total of 4267 and 2104 transfection cliffs are screened from Hela and RAW 264.7 cell lines, respectively, constituting 68$\%$ and 81$\%$ of all data, indicating the widespread occurrence of transfection cliffs. Our findings is that atoms with high attention scores calculated by the mol-attention mechanism block corresponded to key atoms in transfection cliff pairs. This finding indicates that 
significant differences in transfection efficiency caused by subtle structural changes. To test the generalization capability of TransMA, the results demonstrate that TransMA's predicted values maintain a consistent ranking with the actual transfection efficiency values without training on the external dataset.

In this study, our primary contributions can be summarized as follows:

\begin{enumerate}
\item We propose an explainable LNPs transfection efficiency prediction model, named TransMA, which adopts a multi-modal molecular structure fusion architecture. The fine-grained atomic spatial relationship extractor named molecule 3D Transformer captures molecular 3D geometric features, while the coarse-grained atomic sequence extractor named molecule Mamba captures 1D SMILES representation molecular features.
\item TransMA achieves state-of-the-art performance on the current largest LNPs dataset, showing significant improvement across various metrics compared to previous methods.
\item TransMA demonstrates strong interpretability. We construct 4267 and 2104 pairs of transfection cliffs in Hela and RAW 264.7 cell lines to validate the model's identification of key molecular structures. While validating TransMA's prediction results, we find that it can detect substantial transfection efficiency differences caused by minor structural variations (Figure~\ref{figure7}).
\item We collect independent external test data from several sources. TransMA's predicted values consistently maintain the same order as the actual transfection efficiencies, highlighting its robust generalization capacity.
\end{enumerate}
\section{Materials and methods}
\subsection{The method of TransMA to predict LNPs transfection efficiency}
\begin{figure*}[!t]
	\centering
    \includegraphics[width=15cm]  {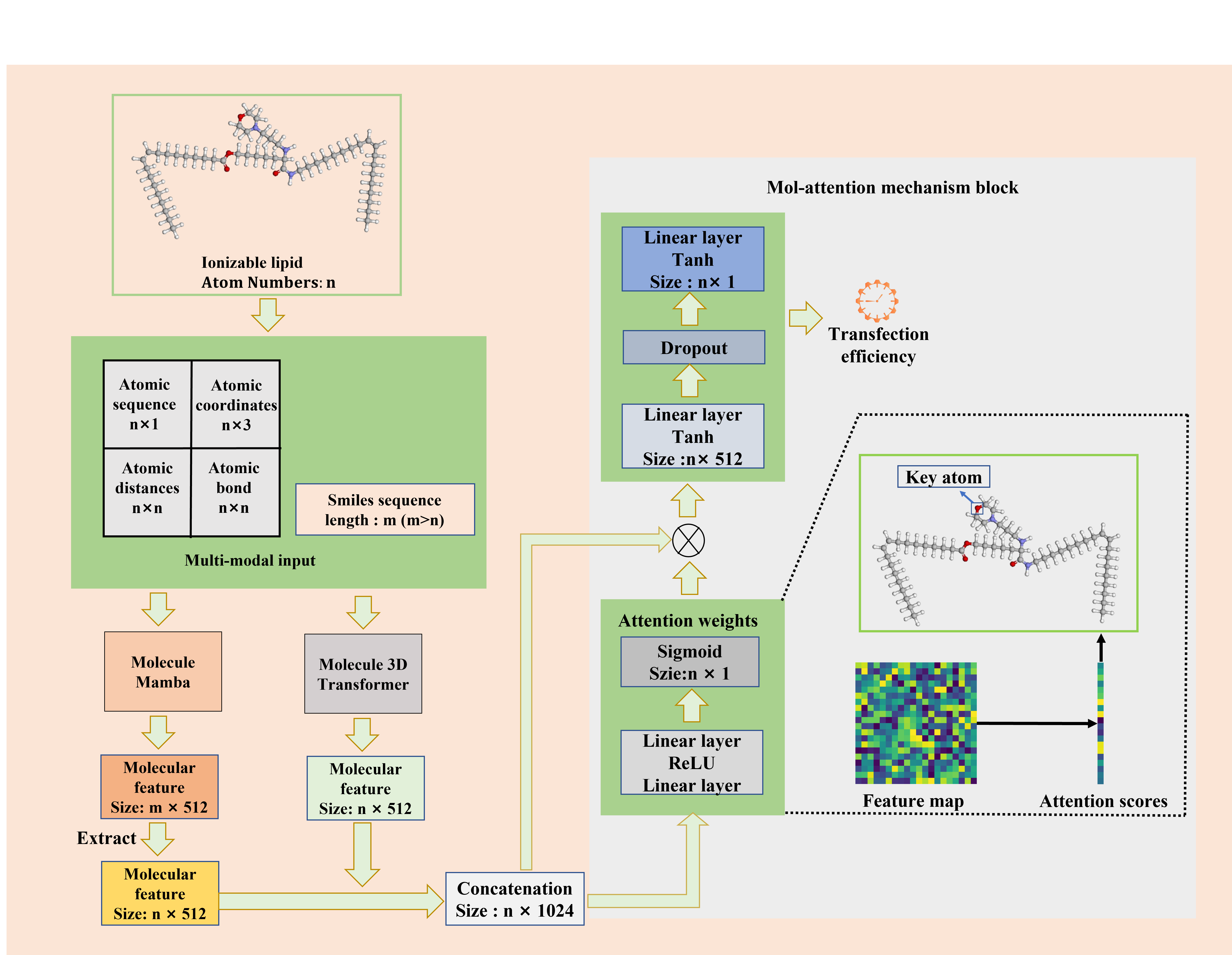}
	\caption{The model takes as input multimodal structural information of ionizable lipids, including three-dimensional structural details: atomic type sequences, three-dimensional coordinates, distance matrices, bond type matrices, and one-dimensional SMILES representation molecular structural information. These two types of structural information undergo feature extraction separately using self-attention mechanism models and spatial state models. The resulting features are fused and visualized through a mol-attention mechanism to reveal the atoms that have a significant impact on transfection efficiency prediction.}
	\label{figure1}
\end{figure*}

Deep learning is a powerful tool for accelerating research in the molecular field\cite{52,53}. We develop an explainable deep learning approach to predict the transfection efficiency of LNPs. This model not only achieves high-precision predictions but also elucidates the key molecular structures that the model focuses on. The overall architecture of TransMA is illustrated in figure~\ref{figure1}. The TransMA framework consists of three components: molecule 3D Transformer, molecule Mamba, and mol-attention mechanism block. 
\subsubsection{Molecule 3D Transformer}
The molecule 3D Transformer is employed to extract the three-dimensional structural information features of molecules. These three-dimensional structural details include the atomic type sequences, atomic three-dimensional coordinates, Euclidean distance matrices between atoms, and matrices representing the types of bonds between atoms in ionizable lipid molecules. TransLNP adopts a pre-training followed by fine-tuning training approach shown in figure~\ref{figure2}A. Despite the scarcity of labeled data with LNPs properties in the LNPs field, we can leverage various molecular structure information from other small molecules to train the model. The pre-training task utilizes masked language modeling: randomly masking 15$\%$ of the atoms and introducing mask tokens. Subsequently, an embedding layer is used to map the atomic sequences to representations of atomic types, providing each atom with semantic information. Equation~\ref{eq0} illustrates the process of self-attention mechanism addressing interactions between atoms:
\begin{equation}
attention = \text{softmax}(\frac{QK}{\sqrt{d}} + bias)V,\label{eq0}
\end{equation}
The masked atomic types are linearly represented as Queries (Q), Keys (K), and Values (V) in the self-attention module. $\sqrt{d}$ represents the dimensions of vectors Q, K, and V. The attention bias in the self-attention mechanism is based on representations of atomic distances and bond types.

After considering the relationships between atoms, uniform noise is introduced and added to the 3D coordinates of the molecule. This process of coordinate updating effectively incorporates the noise into the molecular representation. In the pre-training phase, atomic type heads, atomic coordinate heads, and atomic pairwise Euclidean distance heads are employed to predict masked atomic types, coordinates, and relative distances, respectively. Smooth L1 is used as the loss function for predicting Euclidean distances and masked atomic coordinates, while cross-entropy is used for predicting atomic types. During fine-tuning, the pre-trained model is loaded, following the data processing procedures from pre-training. Finally, molecular prediction heads are used for transfection efficiency prediction. 

\subsubsection{Molecule Mamba}
Mamba is a linear-time sequence modeling method based on selective state spaces\cite{25}. Variants of Mamba have demonstrated outstanding performance in natural language processing, image processing, remote sensing, and audio domains\cite{26,27,28,29}. State Space Sequence Models (SSMs)\cite{30} form the theoretical foundation of Mamba, capable of mapping a one-dimensional function or sequence $u(t)$ to $y(t) \in \mathbb{R}$ through a hidden state $x(t)$. This mapping can be represented by the following linear ordinary differential equation shown in equation~\ref{eq1}: 
\begin{equation}
\dot{x}(t) = Ax(t) + Bu(t), \quad y(t) = Cx(t), \label{eq1}
\end{equation}
where the state matrix $A \in \mathbb{R^{N \times N}}$ serves as the evolution parameter and $B \in \mathbb{R^{N \times 1}}$, $C \in \mathbb{R^{1 \times N}}$ act as projection parameters, representing the implicit latent state.

\begin{figure*}[!t]
	\centering
    \includegraphics[width=15cm]  {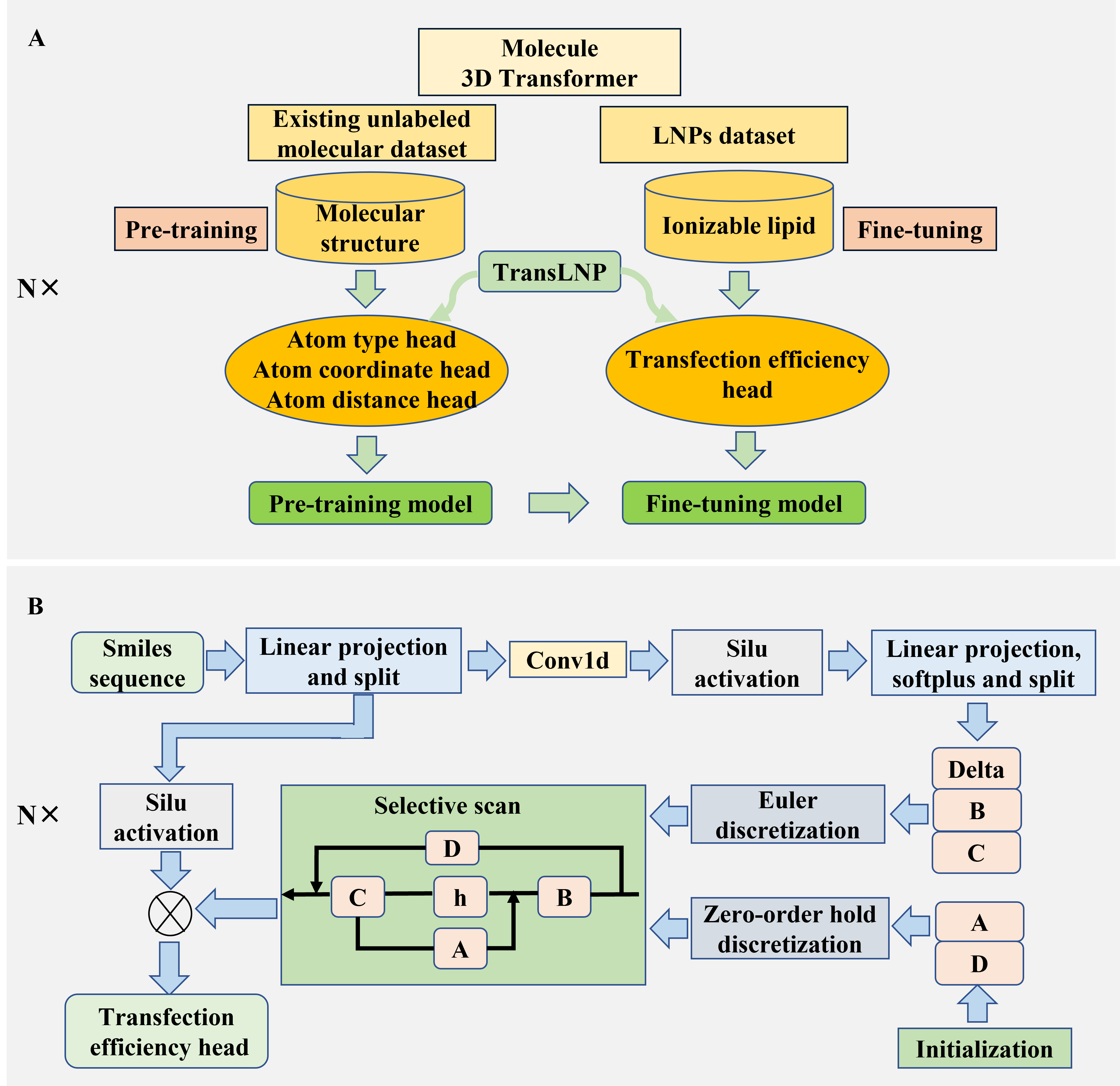}
	\caption{\textbf{Left:}The architecture of the molecule 3D Transformer adopts a pre-training and fine-tuning approach. The pre-training task involves masked language modeling and noise coordinate prediction. Fine-tuning is conducted using molecular prediction heads for transfection efficiency prediction. \textbf{Right:} The input of Molecule Mamba is the SMILES representation of ionizable lipids. Through the selective scan operation and the discretized SSM state equation, high-dimensional features of the molecule are extracted. }
	\label{figure2}
\end{figure*}

Due to its combination of recurrent neural networks (RNNs) and convolutional neural networks (CNNs), Mamba has advantages in handling long sequence data. The SMILES representation of ionizable lipids can be viewed as a long sequence for processing. Therefore, we first apply Mamba to the field of molecular property prediction, using one-dimensional SMILES representation of ionizable lipids as sequence inputs. Figure~\ref{figure2}B illustrates the overall process of Molecule Mamba. Molecule Mamba involves four parameters $(Delta(\Delta), A, B, C)$. Firstly, the feature dimension of the  SMILES representation molecule sequence is doubled through linear projection. After the split operation, the SMILES representation molecule is represented in two parts. One part undergoes one-dimensional convolution and Sliu activation function to extract molecular features. After linear projection and split operation, parameters (Delta, B, C) are obtained. Delta is updated through softplus. At the same time, parameters (A, D) are initialized, where A and D are independent of the input. $A$ is discretized using zero-order hold (ZOH) discretization, while $B$ is discretized using a simplified Euler discretization shown in equation~\ref{eq2}. 
\begin{equation}
\overline{A} = \exp(\Delta A), \overline{B} = x + \Delta  \cdot f(x, u)
, \label{eq2}
\end{equation}
Combining equation~\ref{eq1} with the scanning operation, the sequential processing of elements in the SMILES representation molecule involves updating atoms at each time step based on the cumulative effect of the previous atom and the current input, effectively propagating information across the entire sequence. Equation~\ref{eq3} describes this process. Finally, the output is the multiplication of another portion of the input with the output of the scanning operation.
\begin{equation}
\dot{x}(t) = \overline{A}x(t) + \overline{B}u(t), \quad y(t) = Cx(t), \label{eq3}
\end{equation}
\subsubsection{Mol-attention mechanism block}
The purpose of the mol-attention mechanism block is to integrate the output features of the molecule 3D Transformer and molecule Mamba, and introduce an attention mechanism to explain the model's attention to each atom. The output feature $ z_1 $ of the molecule 3D Transformer has a size of $ [n,512] $, where $ n $ represents the number of atoms in the ionizable lipid. Meanwhile, the output feature $ z_2 $ of molecule Mamba has a size of $ [m,512] $, where $ m $ denotes the sequence length of the SMILES representation molecule. Apart from containing atomic features, $ z_2 $ also includes features of characters in the SMILES representation molecule. Before concatenating the features, all atomic features $ z_2' $ need to be extracted from $ z_2 $ to match the size of $ z_1 $. Therefore each atomic feature in $ z_2' $ corresponds to one atomic feature in $ z_1 $, ensuring correct alignment during feature fusion. Like attention mechanisms in the image domain\cite{32,33,34}, the mol-attention mechanism block aims to weight each atom, compressing the fused features containing global information of size $ n \times 1024 $ directly into a $ n \times 1 $ feature vector $ z $. The first fully connected layer $W_1$ compresses the 1024 concatenated features into $ 1024/r $ channels to reduce computational complexity. Following a ReLU non-linear activation layer $\delta$, the second fully connected layer $W_2$ produces weights $attention\ scores $ through a Sigmoid activation $\sigma$. The dimension of $attention\ scores$ obtained is $ n \times 1 $, which is shown as equation~\ref{eq4}:
\begin{equation}
z = \text{concat}(z_1, z_2'), attention\ scores = \sigma(g(z,W))=\sigma(W_2 \delta (W_1z)), \label{eq4}
\end{equation}
where $ r $ denotes the compression ratio. Finally, the scale operation and a regression layer are applied to predict the transfection efficiency, which is shown as equation~\ref{eq44}:
\begin{equation}
transfection\ efficiency =\tanh (W_2(\tanh( W_1(z \cdot attention\ scores)))), \label{eq44}
\end{equation}
where $\tanh$ denotes the tanh activation function.
\subsection{Loss function}
The loss function of the TransMA model combines the mean squared error (MSE) loss function and the triplet loss function, constructing a hybrid loss function. Equation~\ref{eq5} is shown as: 
\begin{equation}
hybrid\_loss = \frac{1}{N} \sum_{i=1}^{N} (y_i - \hat{y}_i)^2 + \beta \times \mathcal{L}_{\text{triplet}}(z_1, z_2'), \label{eq5}
\end{equation}
where $\beta$ is a weight parameter, $N$ is the number of samples, $y_i$ and $\hat{y}_i$ respectively represent the true values and the predicted values, and the definitions of $z_1$ and $z_2'$ are consistent with equation~\ref{eq4}.
Equation~\ref{eq6} describes the triplet loss:
\begin{equation}
\begin{split}
\mathcal{L}_{\text{triplet}} = \frac{1}{\text{num\_positive\_losses} + \epsilon} \sum_{i=1}^{2batch} \sum_{j=1}^{2batch} \sum_{k=1}^{2batch}\\
\max(d(z_{1i}, z_{1j}) - d(z_{1i}, z_{2k}) + \text{margin}, 0) \cdot \text{mask}_{ijk},\label{eq6}
\end{split}
\end{equation}
where $batch$ represents the batch size, $z_{1i}$ represents the feature extracted from the molecule 3D Transformer for the $i$th sample, $z_{2k}'$ denotes the feature extracted from the molecule Mamba for the $k$th sample, $d(\cdot)$ denotes the Euclidean distance, $\text{num\_positive\_losses}$ is the number of positive losses, $\epsilon$ is a small number used for numerical stability,  $margin$ is a hyperparameter controlling the margin size for effective triplets. and $\text{mask}_{ijk}$ is a triplet mask used to filter out invalid triplets. 

The Triplet Loss first computes the Euclidean distance between $z_1$ and $z_2$. For each sample, it constructs all possible triplets and computes the loss value for each triplet. This loss value is calculated as the difference between the distance from the anchor sample to the positive sample and the distance from the anchor sample to the negative sample, with a margin added, and then taking the maximum value. Subsequently, invalid triplets are filtered out using a triplet mask, and the average loss value is computed as the final result of the triplet Loss. Triplet loss maps the molecular 3D structural features extracted by the molecule 3D Transformer and molecule Mamba, along with the one-dimensional sequence features, to a unified embedding space. By learning the similarity of molecules in this embedding space, the model can better generalize to unseen molecules.
\section{Experimental Section}
\subsection{Dataset}
The dataset for TransMA predicting LNPs transfection efficiency comprises two main components. One part is utilized for pretraining the molecule 3D Transformer, while the other part is used for training the molecule Mamba and fine-tuning the molecule 3D Transformer. The dataset used for pretraining the molecule 3D Transformer is sourced from Unimol\cite{35}, which comprises 19 million diverse molecular structures\footnote{\url{https://bioos-hermite-beijing.tos-cn-beijing.volces.com/unimol_data/pretrain/ligands.tar.gz}}. The dataset utilized for training the molecule Mamba and fine-tuning the molecule 3D Transformer is obtained from AGILE\footnote{\url{https://github.com/bowang-lab/AGILE}}, consisting of 1200 ionizable lipid SMILES representation molecules along with their corresponding transfection efficiency data in Hela and RAW 264.7 cell lines. 

Ionizable lipids comprises 20 head groups, 12 carbon chains with ester bonds, and 5 carbon chains with isocyanide head groups. High-throughput synthesis of LNPs entails mixing an aqueous phase containing mRNA and an ethanol phase containing lipids using the OT-2 pipetting robot. The aqueous phase, prepared in pH 4.0 10 mM sodium citrate buffer, contains firefly luciferase mRNA, Cre recombinase mRNA, or EGFP-mRNA. Meanwhile, the ethanol phase comprises a mixture of 1200 ionizable lipids and fixed proportions of helper phospholipids (DOTAP, DOPE, cholesterol, and C14-PEG 2000), with a lipid-to-mRNA weight ratio of 10:1. Transfection efficiency data were obtained by applying these 1200 mRNA-LNPs to Hela and RAW 264.7 cell lines for in vitro transfection experiments.
\subsection{Experimental processing}
The tasks of molecule 3D Transformer pretraining include using random positions as corrupted input 3D positions and training the model to predict the correct positions, employing different heads to predict distances between corrupted atom pairs, the correct coordinates of corrupted atoms, and to mask and predict the atom types of corrupted atoms. Atomic type prediction employs the cross-entropy loss function with a weight of 1. Prediction related to atomic coordinates and interatomic distances utilizes the smooth L1 loss function with weights set to 5 and 10 respectively. Pretraining parameters are set as follows: model layers are 15, batch size is 128, atom types are 30, learning rate is adjusted using linear decay, and the optimizer is Adam with $\epsilon$. Parameters for molecule 3D Transformer fine-tuning are as follows: model layers are 16, batch size is 4, epochs are set to 200, and the initial learning rate is set to 1e-5.

Molecule Mamba first tokenizes 1200 ionizable lipid SMILES molecular representation. The tokenizer used is ChemBERTa-77M-MTR from Deepchem. The purpose of tokenization is to segment SMILES molecule repren into tokens and convert these tokens into numerical representations that the model can process. The parameters of the Molecule Mamba model are set as follows: the feature dimension is 512, the number of layers is 2, the vocabulary size is 100, and the training process parameters are the same as molecule 3D Transformer fine-tuning. In TransMA's hybrid loss function, the parameter $\beta$ is set to 6 during the scaffpld data splitting and 3 during the cliff data splitting.
\subsection{Comparison with representative deep learning-based molecular property prediction models}
To demonstrate the superiority of the proposed method, compare the predictive transfection efficiency accuracy of TransMA with advanced molecular property prediction models based on graph convolutional neural networks and Transformer on the LNPs dataset. The compared five models include AGILE, large-scale self-supervised pretraining for molecular property prediction model named ChemBERTa\cite{36}, self-supervised graph neural network framework named MolCLR\cite{37}, and geometry enhanced molecular representation learning model named GEM\cite{38}, TransLNP.

\begin{figure*}[!t]
	\centering
    \includegraphics[width=17cm]  {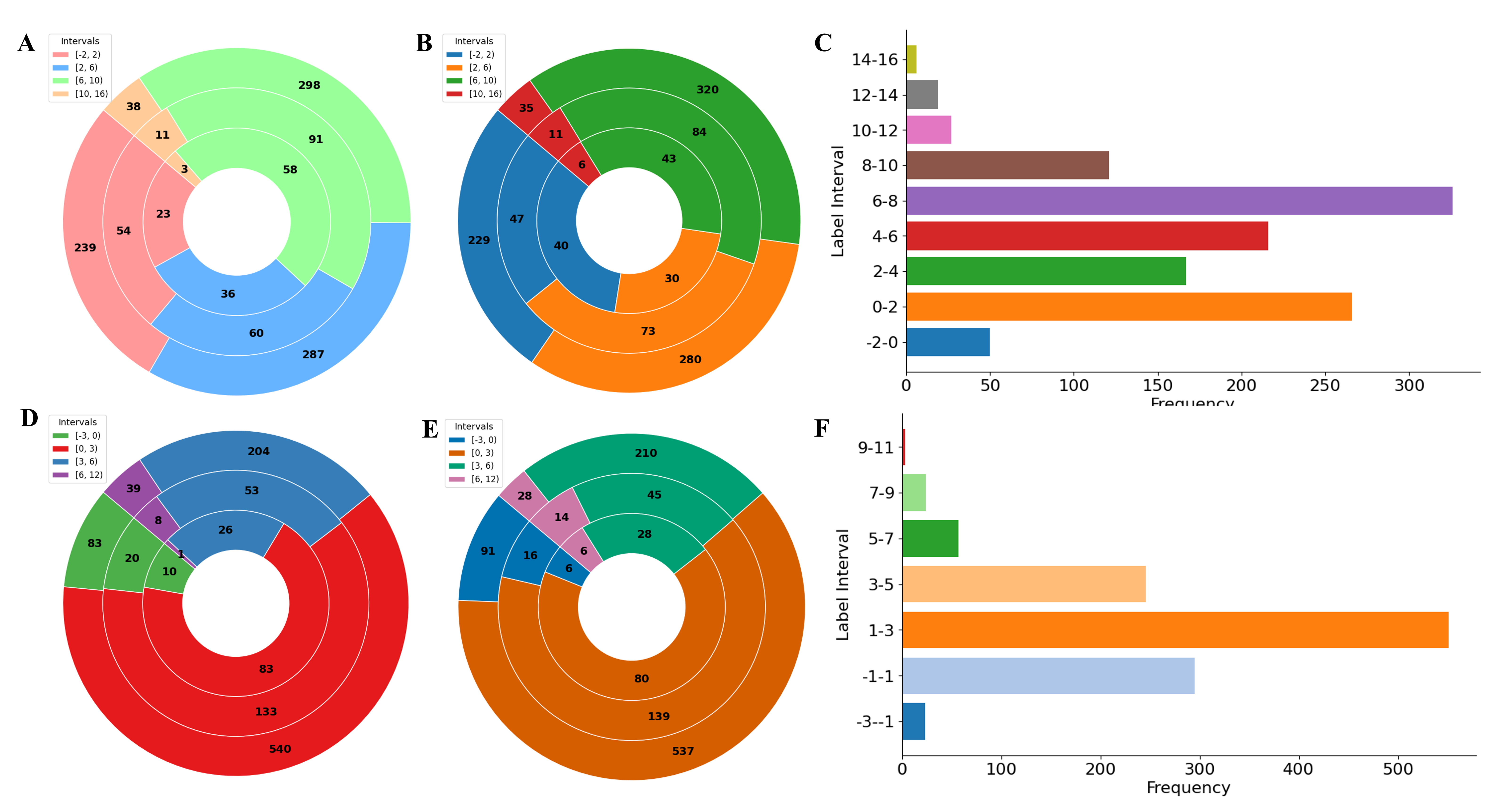}
	\caption{
The distribution of transfection efficiency dataset under two data splitting methods in Hela and RAW 264.7 cell lines is as follows: the outer ring of the pie chart represents the training set distribution, the middle ring represents the validation set distribution, and the inner ring represents the test set distribution.(A) The distribution of transfection efficiency dataset under the scaffold-based train-val-test split in Hela cell line . (B) The distribution of transfection efficiency dataset under the cliff-based train-val-test split in Hela cell line. (C) The distribution of transfection efficiency dataset in Hela cell line. (D) The distribution of transfection efficiency dataset under the scaffold-based train-val-test split in RAW 264.7 cell line. (E) The distribution of transfection efficiency dataset under the cliff-based train-val-test split in RAW 264.7 cell line. (F) The distribution of transfection efficiency dataset in RAW 264.7 cell line.}
	\label{figure3}
\end{figure*}

The LNPs dataset includes 1200 ionizable lipid SMILES molecules and transfection efficiency data obtained from Hela and RAW 264.7 cell lines. In model comparison, the LNPs dataset is divided into cliff\cite{39} and scaffold data splitting approaches. Figure~\ref{figure3} represents LNPs dataset distributions under two data splitting methods in Hela and RAW 264.7 cell lines. Scaffold splitting divides LNPs dataset into training set, validation set and testing set as shown in figure~\ref{figure3}A and figure~\ref{figure3}D. Cliff splitting utilizes molecular ECFP descriptors to represent their structures, employing algorithms like spectral clustering to partition molecules into five clusters. Cliff splitting divides LNPs dataset into training set, validation set and testing set as shown in figure~\ref{figure3}B and figure~\ref{figure3}E. For each cluster, a stratified sampling strategy is employed to allocate 10$\%$ of the molecules to the testing set and other molecules to the training set and validation set in an 8:2 ratio. Cliff splitting ensures that training and test sets contain the proportion of cliff molecule pairs in the training and testing sets while preserving information about molecular structural similarity. Scaffold splitting involves grouping molecules based on their core structural scaffolds, ensuring that both training and test sets contain molecules from diverse scaffold structures\cite{40}. Cliff splitting considers the similarity of data, while scaffold splitting considers the diversity of data. Therefore, combining both cliff and scaffold data splitting methods can comprehensively validate the model's generalization capability.

\begin{figure*}[!h]
	\centering
    \includegraphics[width=18cm]  {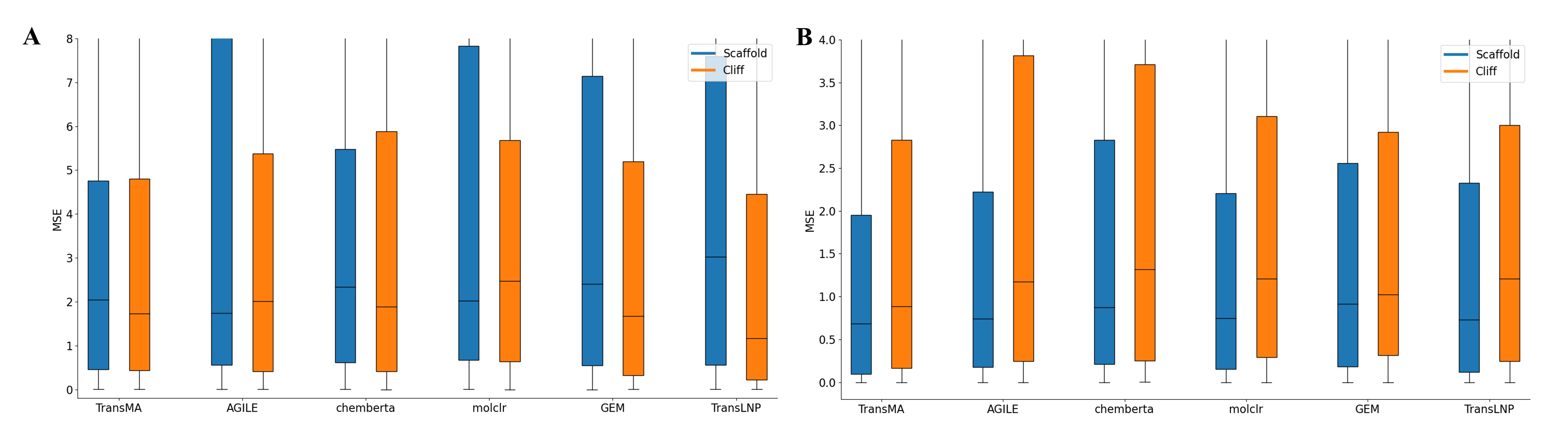}
	\caption{
Comparison of TransMA performance in predicting transfection efficiency with five different models under the scaffold and cliff data splitting methods for Hela and RAW 264.7 cell lines. (A) The box plots comparing TransMA with five models in Hela cell line. (B) The box plots comparing TransMA with five models in RAW 264.7 cell line.}
	\label{figure4}
\end{figure*}

\begin{figure*}[!t]
	\centering
    \includegraphics[width=17cm]  {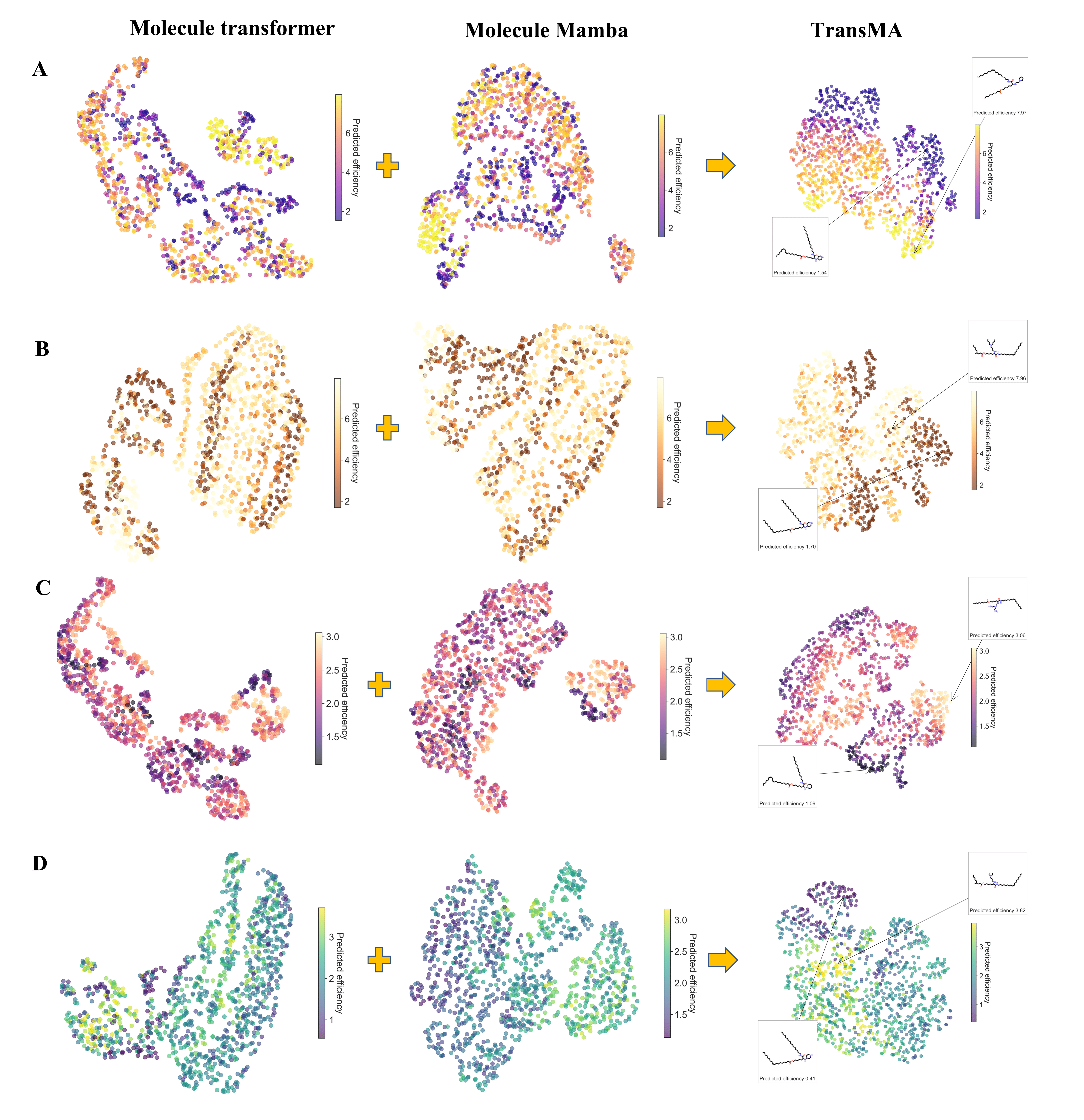}
	\caption{
TransMA multi-modal feature fusion process based on molecule 3D Transformer and molecule Mamba under the scaffold and cliff data splitting methods in Hela and RAW 264.7 cell lines. (A)The UMAP plot of the feature fusion process under scaffold data splitting in Hela cell line. (B)The UMAP plot of the feature fusion process under cliff data splitting in Hela cell line. (C)The UMAP plot of the feature fusion process under scaffold data splitting in RAW 264.7 cell lines. (D)The UMAP plot of the feature fusion process under scaffold data splitting in RAW 264.7 cell line.}
	\label{figure5}
\end{figure*}

Table~\ref{table1} presents the results of TransMA and the compared models in predicting transfection efficiency on the LNPs dataset. Evaluation metrics for assessing prediction performance include mean squared error (MSE), mean absolute error (MAE), the coefficient of determination R$^2$, and the Pearson correlation coefficient (PCC). The results demonstrate that TransMA exhibits superior performance compared to the five other models. Figure~\ref{figure4} shows the boxplots of mean squared error (MSE) for six models in Hela and RAW 264.7 cell lines. It can be observed that TransMA exhibits the lowest MSE in predicting transfection efficiency on both cell lines, indicating the smallest fluctuation in prediction errors. Figure~\ref{figure5} show TransMA's capability to extract multi-modal fused molecular features under the scaffold and cliff data splitting methods in Hela and RAW 264.7 cell lines. It can be observed that the molecular features extracted by molecule 3D Transformer and molecule Mamba exhibit discrete states in different transfection efficiency intervals. After feature fusion by TransMA, the molecular features of TransMA show localization, meaning that the distribution of features for the same transfection efficiency interval is close. Figure~\ref{figure5} indicates that TransMA has successfully established a mapping relationship between the ionizable lipid structure and transfection efficiency.
\begin{table*}[t]
\caption{
The results of predicting LNPs transfection efficiency in both Hela and Raw 264.7 cell lines by six models under scaffold and cliff data splitting methods are presented. Prediction accuracy is assessed using MSE (smaller values indicate better performance), MAE(smaller values indicate better performance), R$^2$ (larger values indicate better performance), and PCC (Pearson correlation coefficient, larger values indicate better performance). "BERTa" refers to the chemBERTa model, while "RAW" refers to Raw 264.7 cells. The best results are highlighted in bold. \protect\label{table1}}
\tabcolsep=0pt%%
\begin{tabular*}{\textwidth}{@{\extracolsep{\fill}}lccccccccccccc@{\extracolsep{\fill}}}
\toprule
\multirow{2}{*}{Splitting} & \multirow{1}{*}{} & \multicolumn{2}{c}{TransMA} & \multicolumn{2}{c}{AGILE} & \multicolumn{2}{c}{BERTa} & \multicolumn{2}{c}{MolCLR} & \multicolumn{2}{c}{GEM} & \multicolumn{2}{c}{TransLNP} \\
\cmidrule{3-14}
& & Hela & RAW & Hela & RAW & Hela & RAW & Hela & RAW & Hela & RAW & Hela & RAW \\
\midrule
\multirow{4}{*}{Scaffold}
& MSE$\downarrow$ & \textbf{3.64} & \textbf{1.63} & 6.38 & 2.07 & 4.38 & 2.11 & 6.18 & 2.00 & 5.89 & 2.24 & 5.58 & 1.88 \\
& MAE$\downarrow$ & \textbf{1.57} & \textbf{0.98} & 1.92 & 1.11 & 1.70 & 1.13 & 1.98 & 1.09 & 1.92 & 1.16 & 1.92 & 1.06 \\
& $R^2$ $\uparrow$ & \textbf{0.49} & \textbf{0.23} & 0.11 & 0.02 & 0.39 & 0.01 & 0.12 & 0.05 & 0.18 & -0.05 & 0.22 & 0.12 \\
& PCC $\uparrow$ & \textbf{0.75} & \textbf{0.54} & 0.46 & 0.24 & 0.63 & 0.36 & 0.42 & 0.25 & 0.48 & 0.24 & 0.47 & 0.37 \\
\midrule
\multirow{4}{*}{Cliff}
& MSE $\downarrow$ & \textbf{4.36} & \textbf{2.88} & 5.94 & 2.90 & 5.34 & 2.98 & 5.96 & 3.07 & 5.51 & 3.17 & 4.47 & 3.06 \\
& MAE $\downarrow$ & \textbf{1.62} & \textbf{1.24} & 1.79 & 1.33 & 1.78 & 1.34 & 1.89 & 1.35 & 1.72 & 1.33 & 1.55 & 1.30 \\
& $R^2$ $\uparrow$ & \textbf{0.61} & \textbf{0.09} & 0.47 & 0.08 & 0.53 & 0.06 & 0.47 & 0.05 & 0.51 & 0.00 & 0.60 & 0.04 \\
& PCC $\uparrow$ & \textbf{0.79} & \textbf{0.34} & 0.69 & 0.30 & 0.73 & 0.27 & 0.69 & 0.29 & 0.74 & 0.29 & 0.78 & 0.27 \\
\bottomrule
\end{tabular*}
\end{table*}

\subsection{Ablation Experiment}
TransMA's ability to extract molecular features comes from both molecule 3D Transformer and molecule Mamba. To demonstrate the superiority of multi-modal feature extraction for molecules, a ablation experiment is designed: comparing the accuracy of predicting transfection efficiency between TransMA, molecule 3D Transformer, and molecule Mamba under scaffold and cliff data splitting methods in Hela and RAW 264.7 cell lines. In the ablation experiment, TransMA, molecule 3D Transformer, and molecule Mamba ensure consistency in training set and testing set, as well as parameter settings. Table~\ref{table2} presents the performance of TransMA, molecule 3D Transformer, and molecule Mamba in predicting transfection efficiency under two different data splitting methods in Hela and RAW 264.7 cell lines. The results demonstrate that TransMA outperforms both molecule 3D Transformer and molecule Mamba in predicting transfection efficiency. The ablation experiment indicates that the fusion of molecule 3D Transformer and molecule Mamba for extracting multi-dimensional molecular structural features is a key factor in achieving superior performance in the task of transfection efficiency prediction.
\begin{table*}[t]
\caption{
The results of predicting LNPs transfection efficiency by TransMA, molecule 3D Transformer, and molecule Mamba under scaffold and cliff data splitting methods in Hela and RAW 264.7 cell lines are presented. The best results are highlighted in bold. \protect\label{table2}}
\tabcolsep=0pt%%
\begin{tabular*}{\textwidth}{@{\extracolsep{\fill}}lcccccccccccc@{\extracolsep{\fill}}}
\toprule%
Splitting &  \multicolumn{6}{c}{Scaffold} &  \multicolumn{6}{c}{cliff} \\
\cmidrule{1-1}\cmidrule{2-7}\cmidrule{8-13}%
\multirow{2}{*}{Method}&\multicolumn{2}{c}{TransMA}&\multicolumn{2}{c}{Molecule Transformer}&\multicolumn{2}{c}{Molecule Mamba} & \multicolumn{2}{c}{TransMA} & \multicolumn{2}{c}{Molecule Transformer}&\multicolumn{2}{c}{Molecule Mamba}\\
\cline{2-3}\cline{4-5}\cline{6-7}\cline{8-9}\cline{10-11}\cline{12-13}%
&Hela&RAW&Hela&RAW&Hela&RAW&Hela&RAW&Hela&RAW&Hela&RAW\\
\midrule
MSE $\downarrow$   &\textbf{3.64}&\textbf{1.63} &5.13&1.77 &6.38&1.90
      &\textbf{4.36}&\textbf{2.88} &4.63&3.02 &5.29&3.00\\
MAE $\downarrow$    &\textbf{1.57}&\textbf{0.98} &1.88&1.02 &2.17&1.07
      &\textbf{1.62}&\textbf{1.24} &\textbf{1.62}&1.29 &1.74&1.30 \\
$R^2$ $\uparrow$  &0.49&\textbf{0.23} &\textbf{0.56}&0.17 &0.11&0.10
      &\textbf{0.61}&\textbf{0.09} &0.59&0.05 &0.53&0.06\\
PCC $\uparrow$    &\textbf{0.75}&\textbf{0.54} &0.28&0.43 &0.39&0.34 
      &\textbf{0.79}&\textbf{0.34} &0.78&0.27 &0.74&0.30 \\
\bottomrule
\end{tabular*}
\end{table*}
\subsection{Analysis of model interpretability}
In this section, the interpretability of the model is analyzed to verify whether TransMA can identify the key atoms in ionizable lipids that affect transfection efficiency. First, identifying the locations of key atoms in ionizable lipids is fundamental for interpretability analysis. Therefore, constructing transfection cliffs is employed to pinpoint key atoms. Concurrently, the mol-attention mechanism block calculates the influence scores of all atoms in ionizable lipids on transfection efficiency. If the high-scoring atoms identified by the mol-attention mechanism align with the key atoms found through the construction of transfection cliffs, it proves that the model can accurately recognize the key atoms in ionizable lipids that impact transfection efficiency.
\subsubsection{Transfection cliffs}
Transfection cliff molecular pairs refer to ionizable lipid molecules that are structurally very similar but have significantly different transfection efficiencies. The phenomenon of transfection cliffs exists in LNP datasets and can be observed through UMAP plots where points that are close in space exhibit large differences in mapped colors. Despite the structural similarity of transfection cliff molecular pairs, there are differences in their atomic compositions, which lead to significant differences in transfection efficiency. Therefore, the atoms that differ between the structures of transfection cliff molecular pairs are considered key atoms affecting transfection efficiency. To quantify the similarity of ionizable lipid molecules, similarity scores based on substructure similarity, scaffold similarity, and SMILES string similarity are constructed\cite{41}. Substructure similarity and scaffold similarity are determined by calculating the Tanimoto coefficient of the Extended Connectivity Fingerprints (ECFP)\cite{42,43} and the Molecular ACCess System (MACCS) keys\cite{44,45} for the molecular graph frameworks, respectively. SMILES string similarity is assessed by calculating the Levenshtein distance\cite{46} between the SMILES string representations of the molecules.
Equation~\ref{eq7} calculates the Tanimoto coefficient shown as:
\begin{equation}
\text{Tanimoto coefficient} = \frac{c}{a + b - c},\label{eq7}
\end{equation}
For substructure similarity, $a$ counts the number of set bits (1s) in molecule A's ECFP fingerprint, $b$ does the same for molecule B, and $c$ counts the number of common set bits in both fingerprints. For scaffold similarity, $a$ counts set bits in molecule A's MACCS keys, $b$ does the same for molecule B, and $c$ counts common set bits in both molecules' MACCS keys. Equation~\ref{eq8} computes SMILES molecular similarity shown as: 
\begin{equation}
\text{SMILES string similarity} = 1 - \frac{d(s_1, s_2)}{\max(|s_1|, |s_2|)}
,\label{eq8}
\end{equation}
where $d(s1, s2)$ is the Levenshtein distance between the SMILES strings of two molecules.

\begin{figure*}[!h]
	\centering
    \includegraphics[width=17cm]  {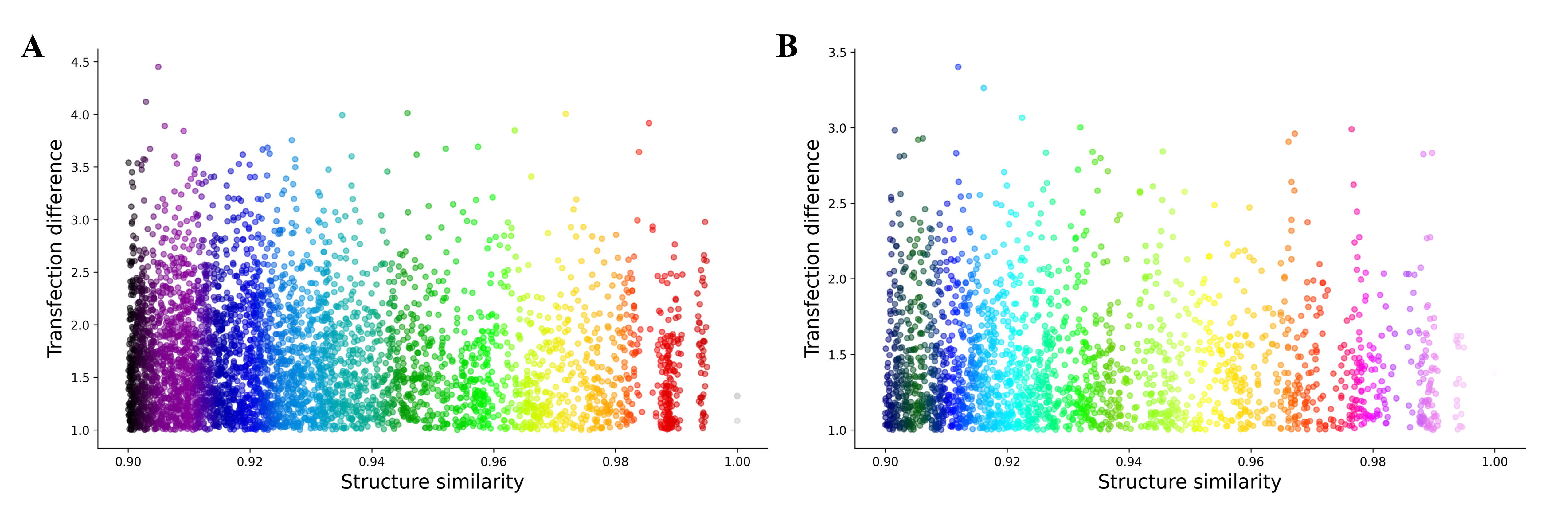}
	\caption{
Scatter plots of structural similarity and transfection difference for transfection cliff pairs in Hela and RAW 264.7 cell lines. (A) Scatter plot of structural similarity and transfection difference for transfection cliff pairs in the Hela cell line. (B) Scatter plot of structural similarity and transfection difference for transfection cliff pairs in the RAW 264.7 cell line. }
	\label{figure6}
\end{figure*}

Similar molecule pairs are defined as pairs with a structural similarity greater than 0.9. Equation~\ref{eq9} describes the structural similarity, which is the average of substructure similarity, scaffold similarity, and SMILES string similarity shown as: 
\begin{equation}
\text{structure similarity} = \frac{subs + scas + smis}{3}
,\label{eq9}
\end{equation}
where $subs$ represents substructure similarity, $scas$ represents scaffold similarity, and $smis$ represents SMILES string similarity.
Since the transfection efficiency in the LNPs dataset has been log2-transformed, similar molecule pairs are identified as transfection cliff pairs when the transfection difference exceeds 1. Equation~\ref{eq10} describes the transfection difference shown as:
\begin{equation}
transfection\ efficiencies = |\log_{10}(2^{{m_2-m1}})| ,\label{eq10}
\end{equation}
where $m_1$ and $m_2$ represent the transfection efficiencies of the two molecules in the similar pair.
 A total of 4267 and 2104 transfection cliff pairs are identified in the Hela and RAW 264.7 cell lines, respectively. Figure~\ref{figure6} shows the scatter plots of structural similarity and transfection difference for transfection cliff pairs in the Hela and RAW 264.7 cell lines. From the figure, it can be seen that when the structural similarity of the molecules is greater than 0.9, the transfection difference is significant, reaching up to a thousand-fold or even ten thousand-fold difference.
\subsubsection{Model interpretability}
\begin{figure*}[!t]
	\centering
    \includegraphics[width=17cm]  {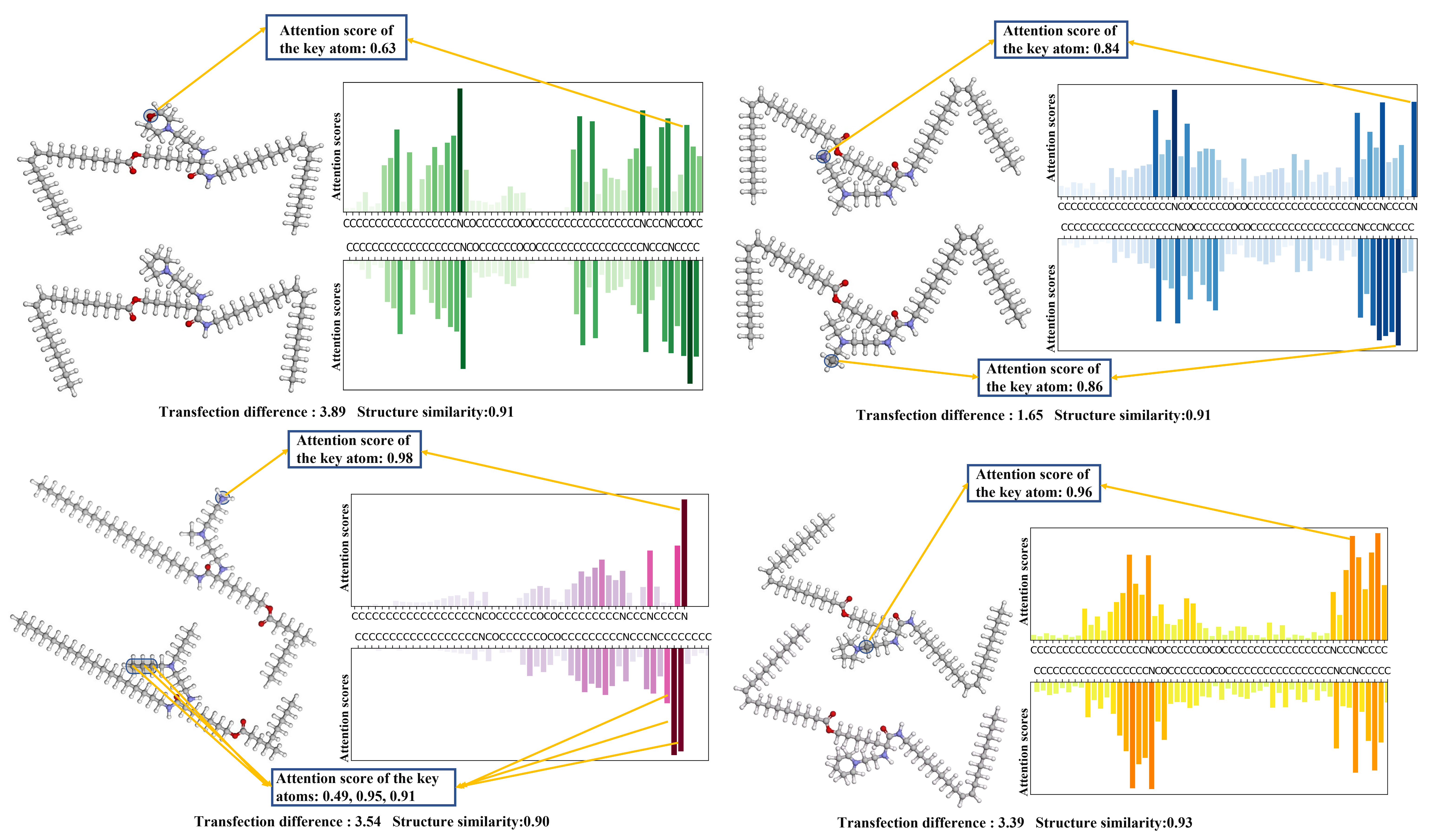}
	\caption{
The scatter plot of attention scores distribution and key atoms display for transfection cliff pairs. The attention score ranges from 0 to 1. The transfection difference value represents the multiplicative difference in transfection efficiency between two molecules, expressed in powers of 10. For instance, a transfection difference of 3.89 implies that the transfection efficiency of one molecule is $10^{3.89}$ times higher than the other. The structural similarity ranges from 0 to 1. }
	\label{figure7}
\end{figure*}

The transfection cliff phenomenon can help identify key atoms in ionizable lipids that affect transfection efficiency. Figure~\ref{figure6} shows that even when the molecular similarity of transfection cliff pairs is close to 1, the transfection difference remains significant. This implies that the one or two differing atoms between the two molecules in the transfection cliff pairs are the cause of the large transfection difference. Therefore, the key atoms are those differing atoms in the transfection cliff pairs. When predicting the transfection efficiency of transfection cliff pairs, the mol-attention mechanism block in TransMA calculates the attention scores for all atoms. Figure~\ref{figure7} shows the attention scores calculated by the mol-attention mechanism block and the key atoms of the transfection cliff pairs. It can be seen that the atoms with the highest attention scores are precisely the key atoms responsible for the differences in the transfection cliff pairs. For example, in figure~\ref{figure7}, the similarity of the transfection cliff pair is 0.91, and the transfection difference is 1.69. This indicates that the transfection efficiency of one molecule is $10^{1.69}$ times higher than the other. The two molecules differ by only one carbon atom (C) and one nitrogen atom (N) in structure. The attention scores corresponding to the differing C and N atoms are 0.84 and 0.86, significantly higher than the attention scores for other atoms. Therefore, TransMA can identify the key atoms affecting transfection efficiency, demonstrating the interpretability of the model.
\subsection{External test}
\begin{figure*}[!h]
	\centering
    \includegraphics[width=17cm]  {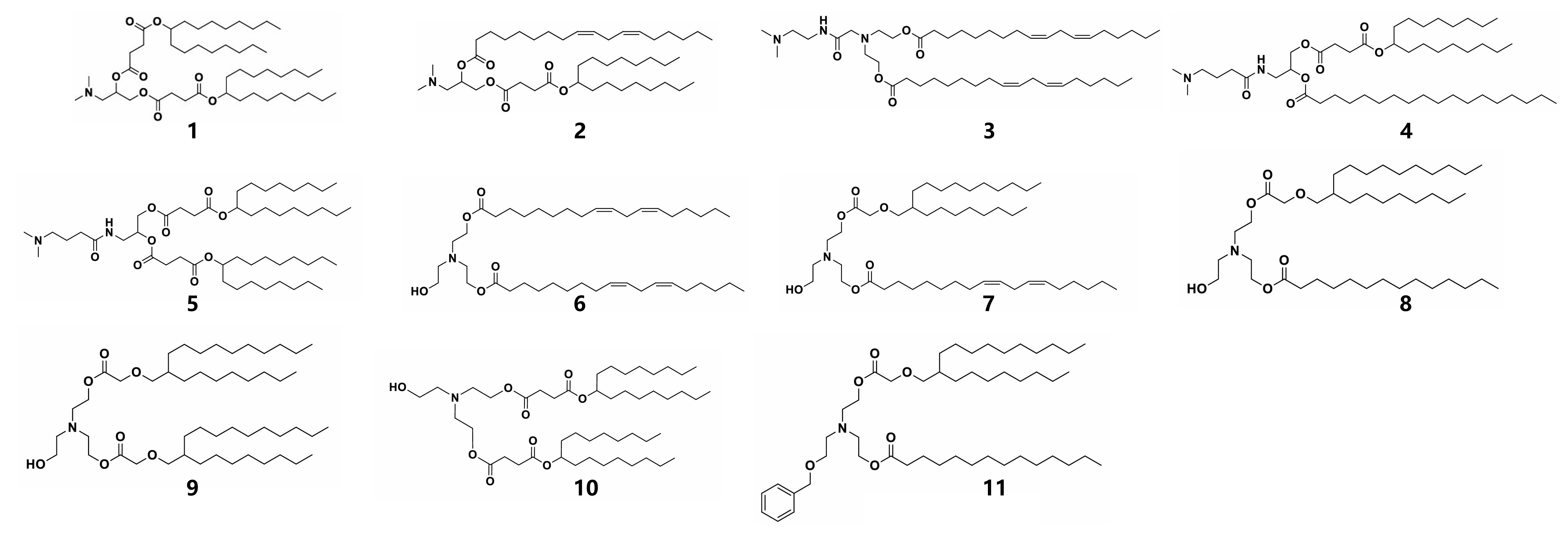}
	\caption{Molecular structures of the 11 ionizable lipids}
	\label{figure8}
\end{figure*}

While TransMA achieves accurate predictions of transfection efficiency on the LNPs dataset, we aim to evaluate its generalization capability by testing it on an external dataset without additional training. We compile an external dataset based on published studies, consisting of 15 LNPs with ionizable lipids as the only variable and their transfection efficiency levels. Among these, 11 ionizable lipids which is from different combinations of 6 head groups, 1 ester linkage, and 5 hydrophobic tails shown in figure~\ref{figure8} and their transfection efficiencies\cite{48} are obtained from Balb/c mice using a fixed LNP formulation. The remaining 4 ionizable lipids and their transfection efficiencies\cite{49,50,51} are derived from different cell lines. Since the TransMA model is also trained on different cell lines, it possesses the capability to predict transfection efficiency across various cell lines.

\begin{figure*}[!h]
	\centering
    \includegraphics[width=17cm]  {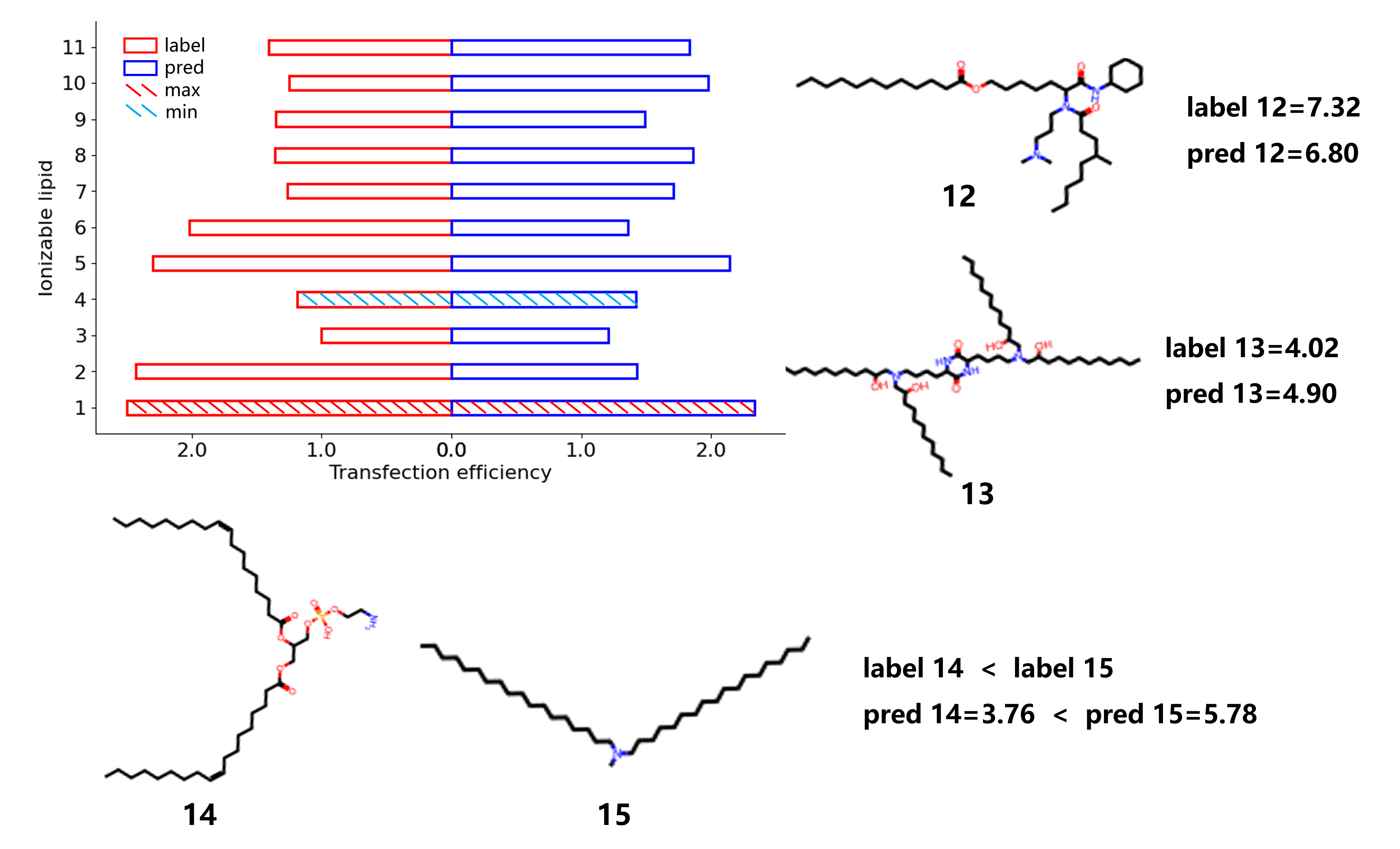}
	\caption{Predicted results for the external dataset without training}
	\label{figure9}
\end{figure*}

Figure~\ref{figure9} shows the results of TransMA directly testing on an external dataset without training. The predicted maximum and minimum transfection efficiencies of lipids 1-11 by TransMA are consistent with the true values, and the order of predicted values is similar to that of true values. For lipids 12 and 13, the prediction error is less than 1. As for lipids 14 and 15, the authors only provided the transfection efficiency of lipid 14 being lower than that of lipid 15 during collection, which is consistent with the prediction result. TransMA achieves good predictions of transfection efficiency even without training on the external dataset, demonstrating its strong generalization capability.
\section{Discussion and conclusion}
In this work, we propose an explainable high-accuracy LNPs transfection efficiency prediction model called TransMA. To achieve high-accuracy predictions, TransMA employs a multi-modal molecular structure fusion architecture, where the substructure molecule 3D Transformer extracts three-dimensional spatial features of the molecule, and molecule mamba extracts one-dimensional molecular features. To achieve model interpretability and feature fusion, we design the mol-attention mechanism block. This block can integrate multi-dimensional features and reveal atom-level structure-transfection relationships based on the molecular channel attention mechanism. Compared with advanced molecular graph convolutional networks and Transformer models, TransMA achieves the highest accuracy in predicting transfection efficiency under the scaffold and cliff data splitting methods in Hela and RAW 264.7 cell lines. Moreover, we introduce transfection cliff pairs. The results demonstrate that the atoms with high attention scores computed by the mol-attention mechanism block correspond to the key atoms in the transfection cliff pairs, showing that TransMA can identify the key atoms affecting transfection efficiency. Additionally, the external test results indicate that even on the untrained external dataset, TransMA's predicted values maintain a consistent order with the actual transfection efficiencies. 

Although TransMA has demonstrated excellent performance in predicting transfection efficiency, there are still limitations in prediction accuracy due to the scarcity of LNPs datasets. Additionally, since ionizable lipids are composed mainly of head and tail groups, different ionizable lipids tend to have high structural similarity, making the occurrence of transfection cliffs more likely. These transfection cliffs increase the difficulty for TransMA to capture the mapping relationship between molecular structure and transfection efficiency. Therefore, future work should focus on addressing the transfection cliff phenomenon to further improve prediction accuracy.

\bibliographystyle{unsrt}
\bibliography{reference}
\end{document}